\documentclass[11pt,a4paper]{article}
\usepackage{authblk}
\usepackage[hyperref]{acl2020}
\usepackage{times}
\usepackage{latexsym}

\usepackage{microtype}

\usepackage{graphicx}
\graphicspath{ {./images/} }
\usepackage{amssymb}
\usepackage{amsmath}
\usepackage{multirow}
\usepackage{makecell}  
\usepackage{tabularx}
\usepackage{url}
\usepackage{pmboxdraw}  

\aclfinalcopy 

\newcommand{\textunderscript}[1]{$_{\text{#1}}$}
\newcommand{\bertbase}[0]{BERT\textunderscript{BASE}}
\newcommand{\bertlarge}[0]{BERT\textunderscript{LARGE}}

\title{Portuguese Named Entity Recognition using BERT-CRF}

\author[1,3]{F\'abio Souza}
\author[2]{Rodrigo Nogueira}
\author[1,3]{Roberto Lotufo}

\affil[1]{University of Campinas \authorcr
   \ \tt{f116735@dac.unicamp.br, lotufo@dca.fee.unicamp.br}}
\affil[2]{New York University \authorcr
  \ {\tt rodrigonogueira@nyu.edu}}
\affil[3]{NeuralMind Intelig\^encia Artificial \authorcr
   \{\tt fabiosouza, roberto\}@neuralmind.ai}

\date{}

\begin{document}
\maketitle
\begin{abstract}
  Recent advances in language representation using neural networks have made it viable to transfer the learned internal states of a trained model to downstream natural language processing tasks, such as named entity recognition (NER) and question answering. It has been shown that the leverage of pre-trained language models improves the overall performance on many tasks and is highly beneficial when labeled data is scarce.
In this work, we train Portuguese BERT models and employ a BERT-CRF architecture to the NER task on the Portuguese language, combining the transfer capabilities of BERT with the structured predictions of CRF. We explore feature-based and fine-tuning training strategies for the BERT model. Our fine-tuning approach obtains new state-of-the-art results on the HAREM I dataset, improving the F1-score by 1 point on the selective scenario (5 NE classes) and by 4 points on the total scenario (10 NE classes).

\end{abstract}

\begin{figure*}[t]
  \centering
  \includegraphics[width=\textwidth]{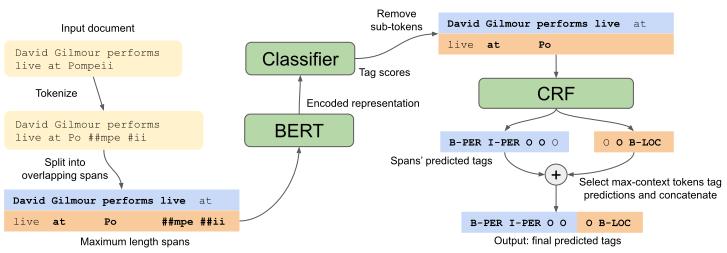}
  \caption{Illustration of the proposed method. Given an input document, the text is tokenized using WordPiece \cite{wu2016google} and the tokenized document is split into overlapping spans of the maximum length using a defined stride (with a stride of 3 in the example). Maximum context tokens of each span are marked in bold. The spans are fed into BERT and then into the classification layer, producing a sequence of tag scores for each span. The sub-token entries (starting with \#\#) are removed from the spans and the remaining tokens are passed to the CRF layer. The maximum context tokens are selected and concatenated to form the final predicted tags.}
  \label{fig:outline}
\end{figure*}

\section{Introduction}

Named entity recognition (NER) is the task of identifying text spans that mention named entities (NEs) and classifying 
them into predefined categories, such as person, organization, location, or any other classes of interest. Despite being conceptually simple, NER is not an easy task. The category of a named entity is highly dependent on textual semantics and its surrounding context. Moreover, there are many definitions of named entity and evaluation criteria, introducing evaluation complications \cite{marrero2013named}.

Current state-of-the-art NER systems employ neural architectures that have been pre-trained on language modeling tasks. Examples of such models are ELMo~\cite{peters2018deep}, OpenAI GPT~\cite{radford2018improving}, BERT~\cite{devlin2018bert}, XLNet~\citep{yang2019xlnet}, RoBERTa~\citep{liu2019roberta}, Albert~\citep{lan2019albert} and T5~\citep{raffel2019t5}. It has been shown that language modeling pre-training significantly improves the performance of many natural language processing tasks and also reduces the amount of labeled data needed for supervised learning \cite{howard2018universal, peters2018deep}.


Applying these recent techniques to the Portuguese language can be highly valuable, given that annotated resources are scarce, but unlabeled text data is abundant. In this work, we assess several neural architectures using BERT (Bidirectional Encoder Representation from Transformers) models to the NER task in Portuguese and compare feature-based and fine-tuning based training strategies. This is the first work to employ BERT models to the NER task in Portuguese. We also discuss the main complications that we face when on existing datasets. With that in mind, we aim to facilitate the reproducibility of this work by making our implementation and models publicly available.\footnote{Code will be available at \href{https://gist.github.com/fabiocapsouza/62c98576d1c826894be2b3ae0993ef53}{\url{https://gist.github.com/fabiocapsouza/62c98576d1c826894be2b3ae0993ef53}}.} \footnote{BERT models available at \href{https://github.com/neuralmind-ai/portuguese-bert}{\url{https://github.com/neuralmind-ai/portuguese-bert}}.}


\section{Related Work}
\label{sect:related_work}

NER systems can be based on handcrafted rules or machine learning approaches. For the Portuguese language, previous works explored machine learning techniques and a few ones applied neural networks models. \citet{amaral2014nerp} created a CRF model using 15 features extracted from the central and surrounding words. \cite{pirovani2018portuguese} combined a CRF model with Local Grammars, following a similar approach.

Starting with \citet{collobert2011natural}, neural network NER systems have become popular due to the minimal feature engineering requirements, which contributes to a higher domain independence \cite{yadav2018survey}. The CharWNN model \cite{santos2015boosting} extended the work of \citet{collobert2011natural} by employing a convolutional layer to extract character-level features from each word. These features were concatenated with pre-trained word embeddings and then used to perform sequential classification.

The CharWNN model \cite{santos2015boosting} extended the work of \citet{collobert2011natural} by employing a convolutional layer to extract character-level features from each word. The LSTM-CRF architecture \cite{lample2016neural} has been commonly used in NER task \cite{castro2018, de2018lener, fernandes2018applying}. The model is composed of two bidirectional LSTM networks that extract and combine character-level and word-level features. A sequential classification is then performed by the CRF layer.

Recent works explored contextual embeddings extracted from language models in conjunction with the LSTM-CRF architecture. \citet{Santos2019MultidomainCE, santos2019assessing} employ Flair Embeddings \citep{akbik2018flair} to extract contextual word embeddings from a bidirectional character-level LM trained on Portuguese corpora. These  embeddings are concatenated with pre-trained word embeddings and fed to a BiLSTM-CRF model. \citet{castro2019elmo} uses ELMo embeddings that are a combination of character-level features extracted by convolutional neural networks and the hidden states of each layer of a bidirectional LM (biLM) composed of a BiLSTM model.

\section{Model}
\label{sect:model}

In this section we describe the model architecture and the training and evaluation procedures for NER.

\subsection{BERT-CRF for NER}

The model architecture is composed of a BERT model with a token-level classifier on top followed by a Linear-Chain CRF. For an input sequence of $n$ tokens, BERT outputs an encoded token sequence with hidden dimension $H$. The classification model projects each token's encoded representation to the tag space, i.e. $ \mathbb{R}^H \mapsto \mathbb{R}^K$, where $K$ is the number of tags and depends on the the number of classes and on the tagging scheme. The output scores $\mathbf{P} \in \mathbb{R}^{n\times K}$ of the classification model are then fed to the CRF layer, whose parameters are a matrix of tag transitions $\mathbf{A} \in \mathbb{R}^{K+2 \times K+2}$. The matrix $\mathbf{A}$ is such that $A_{i,j}$ represents the score of transitioning from tag $i$ to tag $j$. $\mathbf{A}$ includes 2 additional states: start and end of sequence.

As described by \citet{lample2016neural}, 
for an input sequence $ \mathbf{X} = (\mathbf{x}_1, ..., \mathbf{x}_n) $
and a sequence of tag predictions
$ \mathbf{y} = (y_1, ..., y_n), y_i \in \{1, ..., K\} $, the score of the sequence is defined as
$$ s(\mathbf{X}, \mathbf{y}) = \sum_{i=0}^{n} A_{y_i, y_{i+1}} + \sum_{i=1}^{n} P_{i, y_{i}} ,$$
where $y_0$ and $y_{n+1}$ are start and end tags. The model is trained to maximize the log-probability of the correct tag sequence:
\begin{equation}\label{eq:cost}
\log(p(\mathbf{y}|\mathbf{X})) = s(\mathbf{X}, \mathbf{y}) - \log\left(\sum_{\mathbf{\tilde{y}}\in\mathbf{Y_X}}^{} e^{s(\mathbf{X},\mathbf{\tilde{y}})} \right)    
\end{equation}
where $\mathbf{Y_X}$ are all possible tag sequences. The summation in Eq. \ref{eq:cost} is computed using dynamic programming. During evaluation, the most likely sequence is obtained by Viterbi decoding.
Following \citet{devlin2018bert}, we compute predictions and losses only for the first sub-token of each token.

\subsection{Feature-based and Fine-tuning approaches}
We experiment with two transfer learning approaches: \textit{feature-based} and \textit{fine-tuning}. For the feature-based approach, the BERT model weights are kept frozen and only the classifier model and CRF layer are trained. The classifier model consists of a 1-layer BiLSTM with hidden size $d_{LSTM}$ followed by a Linear layer. Instead of using only the last hidden representation layer of BERT, we sum the last 4 layers, following \citet{devlin2018bert}. The resulting architecture resembles the LSTM-CRF model \citet{lample2016neural} but with BERT embeddings.

For the fine-tuning approach, the classifier is a linear layer and all weights, including BERT's, are updated jointly during training. For both approaches, models without the CRF layer are also evaluated. In this case, they are optimized by minimizing the cross entropy loss.

\subsection{Document context and max context evaluation}
\label{sect:max-context}
To take advantage of longer contexts when computing the token representations from BERT, we use document context for input examples instead of sentence context.
Following the approach of \citet{devlin2018bert} on the SQuAD dataset, examples longer than $S$ tokens are broken into spans of length up to $S$ using a stride of $D$ tokens. Each span is used as a separate example during training. During evaluation, however, a single token $T_i$ can be present in $N=\frac{S}{D}$ multiple spans $s_j$, and so may have up to $N$ distinct tag predictions $y_{i,j}$. Each token's final prediction is taken from the span where the token is closer to the central position, that is, the span where it has the most contextual information. Figure \ref{fig:outline} illustrates the evaluation procedure.

\section{Experiments}
\label{sect:experiments}
In this section, we present the experimental setups for BERT pre-trainings and NER training. We present the datasets that are used, the training setups and hyperparameters.

\subsection{BERT pre-trainings}

We train Portuguese BERT models for the two model sizes defined in \citet{devlin2018bert}: BERT Base and BERT Large. The maximum sentence length is set to \hbox{$S=512$} tokens. We train cased models only since capitalization is relevant for NER \cite{castro2018}.

\subsubsection{Vocabulary generation}
A cased Portuguese vocabulary of 30k subword units is generated using SentencePiece \citep{kudo2018sentencepiece} with the BPE algorithm and 200k random Portuguese Wikipedia articles, which is then converted to WordPiece format. Details about SentencePiece to WordPiece conversion can be found in Appendix \ref{appendix:wordpiece-conversion}.

\subsubsection{Pre-training data}

For pre-training data, we use the brWaC corpus \citep{filho2018brwac}, which contains 2.68 billion tokens from 3.53 million documents and is the largest open Portuguese corpus to date. On top of its size, brWaC is composed of whole documents and its methodology ensures high domain diversity and content quality, which are desirable features for BERT pre-training.

 We use only the document body (ignoring the titles) and we apply a single post-processing step on the data to remove \textit{mojibakes}\footnote{Mojibake is a kind of text corruption that occurs when strings are decoded using the incorrect character encoding. For example, the word ``codifica\c{c}\~ao'' becomes ``codifica\~A\S\~A\pounds o'' when encoded in UTF-8 and decoded using ISO-8859-1.} and remnant HTML tags using the \textit{ftfy} library \citep{speer-2019-ftfy}. The final processed corpus has 17.5GB of raw text.

\subsubsection{Pre-training setup}
 The pre-training input sequences are generated with default parameters and use whole work masking (if a word composed of multiple subword units is masked, all of its subword units are masked and have to be predicted in Masked Language Modeling task). The models are trained for 1,000,000 steps. We use a learning rate of 1e-4, learning rate warmup over the first 10,000 steps followed by a linear decay of the learning rate.

For BERT Base models, the weights are initialized with the checkpoint of Multilingual BERT Base. We use a batch size of 128 and sequences of 512 tokens the entire training. This training takes 4 days on a TPUv3-8 instance and performs about 8 epochs over the training data.

For BERT Large, the weights are initialized with the checkpoint of English BERT Large. Since it is a bigger model with longer training time, we follow the instructions of \citet{devlin2018bert} and use sequences of 128 tokens in batches of size 256 for the first 900,000 steps and then sequences of 512 tokens and batch size 128 for the last 100,000 steps. This training takes 7 days on a TPUv3-8 instance and performs about 6 epochs over the training data.

Note that in the calculation of the number of epochs, we are taking into consideration a duplication factor of 10 when generating the input examples. This means that under 10 epochs, the same sentence is seen with different masking and sentence pair in each epoch, which effectively is equal to dynamic example generation.

\subsection{NER experiments}

\subsubsection{NER datasets}

\begin{center}
     \small
     \begin{tabular}{|c|c|c|c|}
      \hline
      \bf{Dataset} &  \bf{Documents} & \bf{Tokens} & \thead{\bf Entities \\ (selective/total)} \\ \hline
      First HAREM  &  129 & 95585 & 4151 / 5017 \\ \hline
      MiniHAREM  & 128  & 64853 & 3018 / 3642 \\
      \hline
     \end{tabular}
     
     \captionof{table}{Dataset and tokenization statistics for the HAREM I corpora. The \textit{Tokens} column refers to whitespace and punctuation tokenization. The \textit{Entities} column comprises the two defined scenarios.}
     \label{table:dataset}
\end{center}


\begin{table*}[ht]

    \small
    \centering\resizebox{1.0\textwidth}{!}{
    \begin{tabular}{|l|c|c|c|c|c|c|}
    
        \hline
        \multirow[c]{2}{*}{\bf Architecture} & \multicolumn{3}{c|}{\bf Total scenario}  & \multicolumn{3}{c|}{ \bf Selective scenario} \\ \cline{2-7} 
                                          & \bf{Prec.}      & \bf{Rec.}    & \bf{F1}    & \bf{Prec.}      & \bf{Rec.}   &  \bf{F1}    \\ \hline
        CharWNN \cite{santos2015boosting} & 67.16           & 63.74        & 65.41      &  73.98          &  68.68      &  71.23   \\ 
        LSTM-CRF \cite{castro2018}        & 72.78           & 68.03        & 70.33      &  78.26          &  74.39      &  76.27    \\ 
        BiLSTM-CRF+FlairBBP \cite{santos2019assessing} & 74.91 & 74.37     & 74.64      &  83.38          &  81.17      &  82.26    \\\hline\hline
        ML-\bertbase-LSTM \dag          & 69.68     & 69.51   & 69.59           &  75.59   &  77.13      &  76.35       \\ 
        ML-\bertbase-LSTM-CRF \dag      & 74.70     & 69.74   & 72.14           &  80.66   &  75.06      &  77.76      \\ 
        ML-\bertbase                    & 72.97     & 73.78   & 73.37           &  77.35   & 79.16       &  78.25       \\ 
        ML-\bertbase-CRF                & 74.82     & 73.49   & 74.15           &  80.10   &  78.78      &  79.44      \\ \hline 
        PT-\bertbase-LSTM \dag          & 75.00     & 73.61   & 74.30           &  79.88   &  80.29      &  80.09       \\ 
        PT-\bertbase-LSTM-CRF \dag      & 78.33     & 73.23   & 75.69           &  84.58   &  78.72      &  81.66      \\ 
        PT-\bertbase                    & 78.36     & 77.62   & 77.98           &  83.22   &  82.85      &  \textbf{83.03}   \\ 
        PT-\bertbase-CRF                & 78.60     & 76.89   & 77.73           &  83.89   &  81.50      &  82.68       \\ \hline
        PT-\bertlarge-LSTM \dag         & 72.96     & 72.05   & 72.50           &  78.13   &  78.93      &  78.53   \\
        PT-\bertlarge-LSTM-CRF \dag     & 77.45     & 72.43   & 74.86           &  83.08   &  77.83      &  80.37   \\
        PT-\bertlarge                   & 78.45     & 77.40   & 77.92           &  83.45   &  83.15      &  \textbf{83.30}   \\
        PT-\bertlarge-CRF               & 80.08     & 77.31   & \textbf{78.67}  &  84.82   &  81.72      &  \textbf{83.24}   \\\hline
    \end{tabular}
    }
    \caption{Comparison of Precision, Recall and F1-scores results on the test set (MiniHAREM). All metrics are calculated using the CoNLL 2003 evaluation script.
    \textbf{Bold} values indicate SOTA results (multiple results bolded if difference within 95\% bootstrap confidence interval). Reported values are the average of multiple runs with different random seeds. \mbox{\dag: feature-based approach.}}
    \label{table:results}
\end{table*}
 
Popular datasets for training and evaluating Portuguese NER are the HAREM Golden Collections (GC) \cite{santos2006, freitas2010second}.
We use the GCs of the First HAREM evaluation contests, which is divided in two subsets: First HAREM and MiniHAREM. Each GC contains manually annotated named entities of 10 classes: Location, Person, Organization, Value, Date, Title, Thing, Event, Abstraction and Other.

Following \citet{santos2015boosting} and \citet{castro2018}, we use the First HAREM as training set and MiniHAREM as test set. The experiments are conducted on two scenarios: a Selective scenario, with 5 entity classes (Person, Organization, Location, Value and Date) and a Total scenario, that considers all 10 classes. Table \ref{table:dataset} contains some dataset statistics.

\subsubsection{HAREM preprocessing}
\label{sect:harem-preprocessing}
The HAREM datasets were annotated taking into consideration vagueness and indeterminacy in text, such as ambiguity in sentences. This way, some text segments contain {\small\textless ALT\textgreater} tags that enclose multiple alternative named entity identification solutions. Additionally, multiple categories may be assigned to a single named entity.

To model NER as a sequence tagging problem, we must select a single truth for each undetermined segment and/or entity. To resolve each {\small\textless ALT\textgreater} tag in the datasets, our approach is to select the alternative that contains the highest number of named entities. In case of ties, the first one is selected. To resolve each named entity that is assigned multiple classes, we simply select the first valid class for the scenario. The dataset preprocessing script is available on GitHub\footnote{\href{https://github.com/fabiocapsouza/harem_preprocessing}{https://github.com/fabiocapsouza/harem\_preprocessing}} and Appendix \ref{appendix:dataset-preprocessing-example} contains an example.

\subsubsection{NER experimental setup}

For NER training, we use 3 distinct BERT models: Multilingual BERT-Base,\footnote{Available at \href{https://github.com/google-research/bert}{https://github.com/google-research/bert}} Portuguese BERT-Base and Portuguese BERT-Large.  We use the IOB2 tagging scheme and a stride of $D=128$ tokens to split the input examples into spans.

The model parameters are divided in two groups with different learning rates: \mbox{5e-5} for BERT model and \mbox{1e-3} for the rest. The numbers of epochs are 100 for BERT-LSTM, 50 for BERT-LSTM-CRF and BERT, and 15 for BERT-CRF. The numbers of epochs are found using a development set comprised of 10\% of the First HAREM training set. We use a batch of size $16$ and the customized Adam optimizer of \citet{devlin2018bert} with weight decay of $0.01$. Similar to pre-training, we use learning rate warmup for the first 10\% of steps and linear decay of the learning rate for the remaining steps.

 To deal with class imbalance, we initialize the bias term of the "O" tag in the linear layer of the classifier with value of $6$ in order to promote a better stability in early training \cite{lin2017focal}. We also use a weight of $0.01$ for "O" tag losses when not using a CRF layer.

For the feature-based approach, we use a \mbox{biLSTM} with 1 layer and hidden dimension of $d_{LSTM} = 100$ units for each direction.

When evaluating, we produce valid predictions by removing all invalid tag transitions for the IOB2 scheme, such as "I-" tags coming directly after "O" tags or after an "I-" tag of a different class. This post-processing step trades off recall for a possibly higher precision.

\section{Results}
\label{sect:results}

The main results of our experiments are presented in Table \ref{table:results}. We compare the performances of our models on the two scenarios (total and selective). All metrics are computed using CoNLL 2003 evaluation script,\footnote{\url{https://www.clips.uantwerpen.be/conll2002/ner/bin/conlleval.txt}} that consists of an entity-level micro F1-score considering only exact matches. 

Our proposed Portuguese BERT-CRF model outperforms the previous state-of-the-art (BiLSTM-CRF+FlairBBP), improving the F1-score by about 1 point on the selective scenario and by 4 points on the total scenario. Interestingly, Flair embeddings outperforms BERT models on English NER \citep{akbik2018flair}. Compared to LSTM-CRF architecure without contextual embeddings, our model outperforms by 8.3 and 7.0 absolute points on F1-score on total and selective scenarios, respectively. 

We also remove the CRF layer to evaluate its contribution. Portuguese BERT (PT-BERT-BASE and PT-BERT-LARGE) also outperforms previous works, even without the enforcement of sequential classification provided by the CRF layer. Models with CRF improves or performs similarly to its simpler variants when comparing the overall F1 scores. We note that in most cases they show higher precision scores but lower recall.

While Portuguese \bertlarge\ models are the highest performers in both scenarios, we observe that they experience performance degradation when used in the feature-based approach, performing worse than their smaller variants but still better than the Multilingual BERT. In addition, it can be seen that \bertlarge\ models do not bring much improvement to the selective scenario when compared to \bertbase\ models. We hypothesize that it is due to the small size of the NER dataset.

The models of the feature-based approach perform significantly worse compared to the ones of the fine-tuning approach. The performance gap is found to be much higher than the reported values for NER on English language \citep{peters2019tune}.

The post-processing step of filtering out invalid transitions for the IOB2 scheme increases the F1-scores by 1.9 and 1.2 points, on average, for the feature-based and fine-tuning approaches, respectively. This step produces a reduction of 0.4 points in the recall, but boosts the precision by 3.5 points, on average.

\section{Conclusion}
\label{sect:conclusion}

We present a new state-of-the-art on the HAREM I corpora by pre-training Portuguese BERT models on a large corpus of unlabeled text and fine-tuning a BERT-CRF model on the Portuguese NER task. Our proposed model outperforms the previous state of the art (BiLSTM-CRF+FlairBBP), even though it was pre-trained on much less data.
Considering the issues regarding preprocessing and dataset decisions that affect evaluation compatibility, we give special attention to reproducibility of our results and we make our code and models publicly available. We hope that by releasing our Portuguese BERT models, others will be able to benchmark and improve the performance of many other NLP tasks in Portuguese. Experiments with more recent and efficient models, such as RoBERTa and T5, are left for future works.

\section{Acknowledgements}

R Lotufo acknowledges the support of the Brazilian government through the CNPq Fellowship ref. 310828/2018-0.

\bibliography{acl2020}
\bibliographystyle{acl_natbib}


\clearpage
\appendix

\section{Appendix}
\label{sec:appendix}






\subsection{SentencePiece to WordPiece conversion}
\label{appendix:wordpiece-conversion}

The generated SentencePiece vocabulary is converted to WordPiece following BERT's tokenization rules. Firstly, all BERT special tokens are inserted ([CLS], [MASK], [SEP], and [UNK]) and all punctuation characters of the Multilingual vocabulary are added to Portuguese vocabulary. Then, since BERT splits the text at whitespace and punctuation prior to applying WordPiece tokenization in the resulting chunks, each SentencePiece token that contains punctuation characters is split at these characters, the punctuations are removed and the resulting subword units are added to the vocabulary \footnote{Splitting at punctuation implies no subword token can contain both punctuation and non-punctuation characters.}. Finally, subword units that do not start with ``\pmboxdrawuni{2581}'' are prefixed with ``\#\#'' and ``\pmboxdrawuni{2581}'' characters are removed from the remaining tokens.

\subsection{HAREM Dataset preprocessing example}
\label{appendix:dataset-preprocessing-example}
An example annotation of HAREM that contains multiple solutions, in XML format, is:

\texttt{<ALT><EM CATEG="PER|ORG">Governo de Cavaco Silva</EM>|<EM CATEG="ORG">Governo</EM> de <EM CATEG="PER" TIPO="INDIVIDUAL">Cavaco Silva </EM></ALT>} \\
where {\small\textless EM\textgreater} is a tag for Named Entity (NE) and ``\texttt{|}'' identifies alternative solutions. This annotation can be equally interpreted as containing the following NEs:
\begin{enumerate}
    \item 1 NE: Person "Governo de Cavaco Silva"
    \item 1 NE: Organization "Governo de Cavaco Silva"
    \item 2 NEs: Organization "Governo" and Person "Cavaco Silva"
    \
\end{enumerate}

The rules described in \ref{sect:harem-preprocessing} would select the third solution in the example above.

\end{document}